\documentclass[letterpaper, 10 pt, conference]{ieeeconf}
\IEEEoverridecommandlockouts                              
\usepackage{amsmath,amsfonts}
\usepackage{algorithmic}
\usepackage{algorithm}
\usepackage{array}
\usepackage[caption=false,font=footnotesize]{subfig}
\usepackage{textcomp}
\usepackage{stfloats}
\usepackage{url}
\usepackage{verbatim}
\usepackage{graphicx}
\usepackage{cite}
\usepackage{svg}
\usepackage{xcolor}
\usepackage{hyperref}
\usepackage{siunitx}
\usepackage{adjustbox}
\usepackage{tikz}\usetikzlibrary{positioning,fit,arrows.meta,backgrounds,calc}
\usetikzlibrary{decorations.pathmorphing}
\usepackage{cleveref}

\tikzset{
    decisionmakingmodule/.style={%
        draw, rounded corners,
        minimum width=5cm,
        minimum height=1cm,
        font={\fontsize{10pt}{12}\sffamily},
        fill=brown,
        text=white,
        },
    module/.style={%
        draw, rounded corners,
        minimum width=5cm,
        minimum height=1cm,
        font={\fontsize{10pt}{12}\sffamily},
        fill=teal,
        text=white
        },
    message/.style={
        draw, rounded corners,
        minimum width=5cm,
        minimum height=1cm,
        font={\fontsize{10pt}{12}\sffamily},
        fill=violet,
        text=white
    },
    hardware/.style={
        draw, rounded corners,
        minimum width=5cm,
        minimum height=1cm,
        font={\fontsize{10pt}{12}\sffamily},
        fill=darkgray,
        text=white
    },
    abstract/.style= {
        draw, rounded corners,
        minimum width=2.5cm,
        minimum height=1cm,
        font={\fontsize{10pt}{12}\sffamily},
    },
    sensorshardware/.style={
        draw, rounded corners,
        minimum width=2.5cm,
        minimum height=1cm,
        font={\fontsize{10pt}{12}\sffamily},
        fill=darkgray,
        text=white
    },
    normaltext/.style={
        font={\fontsize{10pt}{12}\sffamily},
    },
    snake it/.style={decorate, decoration=snake},
    rosnode/.style={
        draw,
        minimum width=5cm,
        minimum height=1cm,
        font={\fontsize{10pt}{12}\sffamily},
        fill=blue,
        text=white,
    }
}
\definecolor{rosnodecolor}{HTML}{CFE2F3}
\definecolor{rostopiccolor}{HTML}{D9D2E9}

\newcommand{\mB}{\mathcal{B}}
\newcommand{\mM}{\mathcal{M}}
\newcommand{\mI}{\mathcal{I}}

\title{\LARGE \bf MRNAV: Multi-Robot Aware Planning and Control Stack for Collision and Deadlock-free  Navigation in Cluttered Environments}

\author{Bask{\i}n \c{S}enba\c{s}lar$^{1}$, Pilar Luiz$^{1}$, Wolfgang H\"onig$^{2}$, and Gaurav S. Sukhatme$^{1}$
\thanks{$^{1}$Bask{\i}n \c{S}enba\c{s}lar, Pilar Luiz, and Gaurav S. Sukhatme are with Viterbi School of Engineering,
        University of Southern California, 3650 McClintock Ave,  Los Angeles, CA  90089, USA.
        {\tt\small {baskin.senbaslar, pluiz, gaurav}@usc.edu}}%
\thanks{$^{2}$Wolfgang H\"onig is with the Department of Electrical Engineering and Computer Science,
Technical University of Berlin, Marchstr. 23, 10587 Berlin, Germany.
        {\tt\small hoenig@tu-berlin.de}}%
}

\begin{document}

\maketitle

\thispagestyle{empty}
\pagestyle{empty}

\begin{abstract}
Multi-robot collision-free and deadlock-free navigation in cluttered environments with static and dynamic obstacles is a fundamental problem for many applications.
We introduce MRNAV, a framework for planning and control to effectively navigate in such environments.
Our design utilizes short, medium, and long horizon decision making modules with qualitatively different properties, and defines the responsibilities of them.
The decision making modules complement each other and provide the effective navigation capability.
MRNAV is the first hierarchical approach combining these three levels of decision making modules and explicitly defining their responsibilities.
We implement our design for simulated multi-quadrotor flight.
In our evaluations, we show that all three modules are required for effective navigation in diverse situations.
We show the long-term executability of our approach in an eight hour long wall time (six hour long simulation time) uninterrupted simulation without collisions or deadlocks.\looseness=-1
\end{abstract}


\section*{Supplemental Videos}
\noindent\textbf{Experiment Recordings:} \href{https://youtu.be/6WC0YCEctoE}{https://youtu.be/6WC0YCEctoE}\\
\textbf{Long Term Execution:} \ \ \href{https://youtu.be/LEMNKBRULg4}{https://youtu.be/LEMNKBRULg4}
\section{Introduction}

Collision-free and deadlock-free navigation of mobile robots is a fundamental problem for many applications including autonomous driving~\cite{campbell2010autonomous}, 
autonomous last-mile delivery~\cite{li2020lastmile}, 
and automated warehouses~\cite{kiva}.
Here, collision-free navigation refers to the property that robots do not collide with each other or obstacles, and deadlock-free navigation refers to the property that robots reach their goal positions whenever there is a collision-free way of doing so, which is a property that is related to completeness.
The complexity of the navigation task in such applications varies: some contain multiple robots, i.e., teammates, some contain no obstacles, while others contain static or dynamic obstacles.\looseness=-1

In this paper, we present MRNAV, a planning and control stack design for collision and deadlock-free navigation of multi-robot teams in cluttered environments, which might contain static and dynamic obstacles.
Our system provides a solution for navigation in all types of environments.
We discuss the design as well as deployment options for the components.
MRNAV has a clear interface to interoperate with perception, localization, mapping, and prediction subsystems that might exist in a full robotic navigation stack.\looseness=-1

The main novelty of our design is utilizing three different navigation decision making components, namely long, medium, and short horizon decision making modules, with qualitatively different collision and deadlock-free navigation capabilities and argue that effective multi-robot navigation requires a harmonious operation of them.
These methods are often seen as alternatives of each other in the literature, in which, only one of them is used for collision avoidance in conventional designs.
We argue that these qualitatively different multi-robot navigation decision making approaches are not alternatives, but complements of each other in a full navigation stack.
MRNAV is the first hierarchical approach combining these three qualitatively different decision making components.\looseness=-1

\begin{figure}[t]
    \centering
    \includegraphics[width=\linewidth]{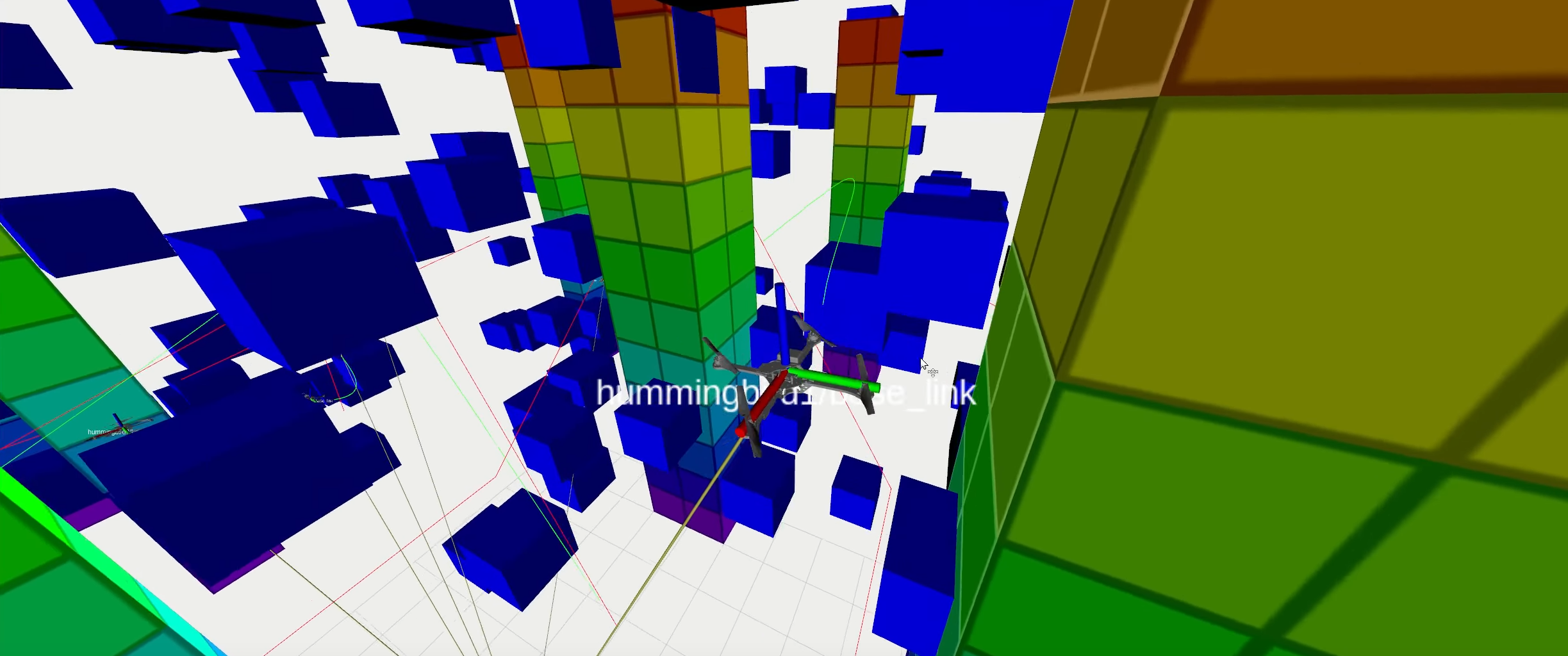}
    \vspace{-0.25in}
    \caption{We implement MRNAV for collision-free multiple quadrotor flight in highly cluttered environments with static and dynamic obstacles.}
    \label{Figure:ClutteredEnvironment}
    \vspace{-0.25in}
\end{figure}

Vital properties of MRNAV include assurance of dynamic feasibility of the actions generated, real-time executability, compliance with communication imperfections (including message drops, delays, and re-orderings, and not depending on communication for collision-free operation), and reasonable liveness of the system, i.e., not being overly conservative for collision-free operation.\looseness=-1

We provide a concrete implementation of our design for navigation of multiple quadrotors in highly cluttered environments with static and dynamic obstacles (Fig.~\ref{Figure:ClutteredEnvironment}).
We run experiments using simulated quadrotors with motor speed control, and experimentally show that MRNAV's three modules are required for effective collision-free and deadlock-free navigation.
In order to demonstrate the long term executability of our implementation, we run an eight hour wall time (six hour simulation time) long uninterrupted simulation with static and dynamic obstacles in which eight quadrotors navigate without collisions or deadlocks.\looseness=-1


\section{Related Work}

The decision making approaches for (multi-)robot navigation are investigated in many domains using different abstractions, assumptions, formalisms, and methods.
We discuss them from the perspective that is important for our system design.
We classify navigation decision making approaches into three categories: long, medium, and short horizon, differing in the planning horizon used to compute an output.\looseness=-1

\subsection{Long Horizon Navigation Decision Making (LH-NDM)}\label{Section:RelatedWork/LongHorizonNavigationDecisionMaking}

LH-NDM approaches are concerned with making decisions for the full navigation task where the planning horizon (distance or duration) may grow to infinity; they are sometimes referred to as global planners.
They can produce trajectories or paths from start to goal states, or they can solve joint task and motion planning (TAMP) problems.
Completeness is one of the central aims during LH-NDM, meaning that the algorithm produces a solution if the navigation problem is feasible, albeit not in real-time. 
Because of this, using LH-NDM approaches with completeness guarantees online in real-time is not viable for the general navigation task.
Dynamic obstacle avoidance is generally not pursued in LH-NDM, since knowing the existence and behaviors dynamic obstacles before navigation, i.e., offline, is not a realistic expectation and incorporating them to decision making is often unnecessary since by the time long horizon plans are executed, dynamic objects may have already moved contrary to their predictions in many navigation applications.\looseness=-1

Many planning algorithms can be used for single-robot LH-NDM, e.g., RRT*~\cite{karaman2010rrtstar} or A*.
Combining planning algorithms with trajectory optimization to ensure the dynamic feasibility of the final plans is possible~\cite{richter2013planning}.
Multi-agent path finding (MAPF)~\cite{stern2019multi} 
is concerned with finding synchronized paths for multiple agents on graphs such that no two agents occupy the same vertex or traverse the same edge in opposite directions at the same step, i.e., they do not collide.
Many optimal~\cite{sharon2015cbs}, bounded suboptimal~\cite{li2021eecbs}, or unbounded suboptimal algorithms~\cite{luna2011push} with cost metrics such as sum-of-costs or makespan are proposed for the MAPF problem.
Generally, robot dynamics are omitted during path planning.
MAPF algorithms typically provide completeness guarantee.
Combining MAPF algorithms with trajectory optimization allows dynamic feasibility of the produced plans~\cite{honig2018quadswarms}, while some approaches do not decouple path planning from robot dynamics~\cite{chen2021scalable}.\looseness=-1

\subsection{Medium Horizon Navigation Decision Making (MH-NDM)}

MH-NDM approaches generate plans for reaching intermediate goals, and are sometimes called local planners.
They generate plans for the medium horizon future of the navigation task, e.g., the next two to ten seconds.
They are typically used in a receding horizon fashion, generating plans that are executed for a short duration followed by re-planning.
The re-planning frequency of MH-NDM approaches typically vary from $\SI{1}{Hz}$ to $\SI{20}{Hz}$.
MH-NDM algorithms are expected to run in real-time because of the receding horizon execution, requiring on-board deployment unless realiable communication maintenance from a centralized computer to the robots is possible.
The real-time planning requirement makes it difficult to have completeness guarantees within the re-planning period. 
On-board MH-NDM allows integration of new information about the environment that may be collected during navigation, making them a viable choice in unknown or partially known environments and tasks with imperfect communication media.
In addition, since their decision making horizon is not long, the robot does not need to maintain the full representation of the environment, allowing on-board local environment representation maintenance.
On-board execution complements with decentralization of planning when there are multiple robots to be controlled, in which each robot computes its own plan, while possibly coordinating with other robots through explicit communication or implicit behavioral assumptions, e.g., assuming that teammates run the same algorithm.
Dynamic feasibility of the generated plans are frequently pursued in MH-NDM.
Generally, the feasibility of most MH-NDM algorithm is not guaranteed, i.e., planning iterations might fail.
In those cases, a robot continues using the plan from the previous planning iteration.\looseness=-1

Model-predictive control (MPC) based multi-robot medium horizon receding horizon planners are proposed, requiring position-only sensing~\cite{zhou2017bvc}, full state sensing~\cite{luis2019dmpc}, full future plan communication~\cite{luis2020dmpc}, and ensuring safety under asynchronous planning and unbounded communication delays~\cite{senbaslar2022async}.
Some planners exploit the differential flatness~\cite{murray1995differential} of the robots, providing static obstacle and teammate avoidance with position-only sensing and no communication~\cite{senbaslar2019rte,senbaslar2023rlss}, static obstacle and teammate avoidance with instantaneously communicated plans~\cite{park2022lsc}, static and non-interactive dynamic obstacle and teammate avoidance with instantaneously communicated plans~\cite{tordesillas2022mader} or sensed positions and velocities of objects~\cite{park2021rsfc}, static, and non-interactive dynamic obstacle avoidance, and asynchronously planning teammate avoidance with bounded communication delays~\cite{kondo2023rmader}, and static and non-interactive or interactive dynamic obstacle\footnote{Interactive dynamic obstacles are obstacles that change their behavior depending on the behavior of other objects.\looseness=-1} avoidance, and asynchronously planning teammate avoidance with unbounded communication delays~\cite{senbaslar2023dream}.
Reasoning about uncertainty of perception~\cite{chen2023rast,senbaslar2023dream,zhu2019chance}, prediction~\cite{tordesillas2022mader,kondo2023rmader,senbaslar2023dream}, system dynamics~\cite{zhu2019chance}, or localization~\cite{chen2023rast,zhu2019chance} in MH-NDM are also investigated.
Guiding MH-NDM modules with desired trajectories is possible~\cite{senbaslar2023rlss,senbaslar2023dream,senbaslar2019rte}, enabling the integration with LH-NDM approaches.\looseness=-1

\subsection{Short Horizon Navigation Decision Making (SH-NDM)}

SH-NDM approaches generate only the next safe actions to execute.
Some maintain the invariance of a safe set~\cite{blanchini1999set}, ensuring that there will always be safe actions to execute in the future.
Since they produce only the next action, they typically run at higher frequency than the medium horizon approaches, e.g., from $\SI{50}{Hz}$ to $\SI{200}{Hz}$ depending on the robot dynamics.
Since their reasoning horizons are short and a high frequency execution is required, they typically run on-board using local environment information to ensure safety.
Robots navigating solely using these approaches tend to deadlock, since they generally do not consider task completion during decision making.
Integrating learned models for reducing deadlocks is proposed, e.g.,~\cite{riviere2020glas, cui2022learning}, but, all existing methods are prone to deadlocks.
Robots navigating with SH-NDM algorithms may cooperate with each other explicitly via communication, or implicitly via behavioral assumptions similar to MH-NDM.\looseness=-1

Computing acceleration commands by making desired acceleration commands safe using safety barrier certificates (SBC) ~\cite{wang2017safety} and computing velocity commands by making desired velocities safe~\cite{van2011reciprocal} are proposed, both of which allowing static and dynamic obstacle and teammate avoidance.
Both approaches are \emph{minimally invasive} algorithms, making desired control actions safe by minimally altering them.
Utilizing SBCs to alter unsafe reference trajectories of quadrotors in a minimally invasive manner is investigated~\cite{li2017flatctrl}.
Minimally invasive algorithms can be guided with reference trajectories, allowing MH-NDM and SH-NDM integration.
Computing safe forces and moments for obstacle avoidance in quadrotors is also possible~\cite{wu2016safety}.
GLAS~\cite{riviere2020glas} combines a learned network that is trained to mimic a long horizon planner~\cite{honig2018quadswarms} with a theoretical safety module, which outputs acceleration or velocity commands.
RL-CBF\cite{cui2022learning} combines an reinforcement learning (RL) trained network with a safety barrier certificates based module to guarantee the collision-free operation.
Artificial potential fields are also applied to multi-robot collision avoidance\cite{min2010potential}.\looseness=-1

\section{System Design \& Implementation}

Our system design includes long, medium, and short horizon decision making modules integrated together.
We discuss major challenges of multi-robot navigation to justify our design, deployment, and the algorithmic choices we make later in~\Cref{Section:ImplementationForQuadrotorFlight} for flight of multiple quadrotors.\looseness=-1

\subsection{Major Design Challenges}

\subsubsection{Limited On-board Compute, Memory and Storage}

We assume that the on-board memory and storage capabilities of the robots are limited, which prohibits them from maintaining global environment representation, but they can maintain local environment representations.
In addition, we assume that the on-board compute capabilities of the robots are also limited.
These prohibit the robots from computing long horizon plans with completeness guarantees on-board.\looseness=-1

\subsubsection{Imperfect Communication}

We embrace that the communication systems are inherently imperfect, and sometimes communication can be lost completely.
This necessitates on-board communication-resilient autonomy.
Robots should be able to ensure collision-free operation without communicating with a centralized system or between each other.
In addition to the collision-free operation, robots should be able to locally resolve deadlocks\footnote{While local deadlock resolution does not have an exact definition as global deadlock resolution, i.e., completeness, we use the term to refer to the ever-increasing ability of the robots to resolve deadlocks using their on-board decision making modules.}, because communication to the centralized system may not be maintained.
Since the robots cannot rely on inter-robot communication for robot-robot collision avoidance, they need sensing and estimation capabilities to assess each other's states.\looseness=-1

\subsubsection{Imperfect Prior Knowledge}

In general, it is impossible to know all of the static or dynamic obstacles a priori, except for the use cases in which the environment is specifically designed or obstacle free.
Generally, some static obstacles are known a priori.
Since dynamic obstacles are transitionary, modelling and predicting them offline is unrealistic. 
This necessitates on-board decision making autonomy as well as local environment representation building.
On-board collision-free navigation and local deadlock resolution capabilities allow robots to handle problems stemming from imperfect prior knowledge. \looseness=-1

\subsubsection{Imperfect Reference Trajectory Tracking}

The control systems of mobile robots are characteristically imperfect, in which local deviations from reference trajectories are expected, and explicitly accounted for with feedback control.
Local deviations can stem from unmodeled physics, disturbances, state estimation errors, or imperfect system identification.
When the robots deviate from their reference trajectories in an unforeseen amount, safety guarantees the trajectories satisfy may become invalid.
Therefore, providing collision avoidance guarantees during reference trajectory computation is not enough for collision-free operation, and the control actions executed by the robot should re-evaluate safety at every control step.\looseness=-1

\subsubsection{Minimal Sensing and Estimation Capabilities}

Sensing and estimating the states of entities produces outputs with errors.
We consider current positions and shapes of objects as first order information, such that robots can use inputs from their sensors to estimate these values directly.
Higher order derivative estimates can be done by using a numerical estimator, which increases the error of the estimation.
For the reliability of the safety properties, the robots should need as few higher order derivatives as possible to enforce safety.
In general, position and shape sensing for static objects, and position, velocity and shape sensing for non-static objects is preferred, as these are minimal representations to describe the current state and the immediate future of these objects.\looseness=-1

\subsubsection{High Reactivity}

Behavior prediction becomes a central problem when there are dynamic obstacles.
In general, these predictions are the best real-time estimations of the behaviors, and may fail to describe the medium to long horizon trajectories of the dynamic obstacles.
When dynamic obstacles move unlike their predictions, the collision-free operation properties may be violated. 
In such situations, the robots should be able to react to unforeseen circumstances swiftly for collision-free operation.\looseness=-1

\subsection{Design}

\begin{figure*}[t]
\centering
\resizebox{\textwidth}{!}{%
\begin{tikzpicture}[]

    \node[decisionmakingmodule] (LongHorizonDecisionMaking) {\shortstack{Long Horizon Navigation \\ Decision Making (LH-NDM)}};

    \node[message, above=0.5cm of LongHorizonDecisionMaking] (PriorStaticObstacles) {\shortstack{Prior Static Obstacles}};
    \node[message, left=0.5cm of PriorStaticObstacles] (PriorTeammateStates) {\shortstack{Prior Robot States}};
    \node[message, left=0.5cm of PriorTeammateStates] (TeammateGoalPositions) {\shortstack{Robot Goal Positions}};
    \node[message, right=0.5cm of PriorStaticObstacles] (TeammateShapes) {\shortstack{Robot Shapes}};
    \node[message, right=0.5cm of TeammateShapes] (TeammateDynamics) {\shortstack{Robot Dynamics}};

    \draw[->] (TeammateGoalPositions)--($(TeammateGoalPositions)+(0.0, -0.5)$)--($(LongHorizonDecisionMaking)+(0.0, 0.5)$);
    \draw[->] (PriorTeammateStates)--($(PriorTeammateStates)+(0.0, -0.5)$)--($(LongHorizonDecisionMaking)+(0.0, 0.5)$);
    \draw[->] (PriorStaticObstacles)--($(PriorStaticObstacles)+(0.0, -0.5)$)--($(LongHorizonDecisionMaking)+(0.0, 0.5)$);
    \draw[->] (TeammateShapes)--($(TeammateShapes)+(0.0, -0.5)$)--($(LongHorizonDecisionMaking)+(0.0, 0.5)$);
    \draw[->] (TeammateDynamics)--($(TeammateDynamics)+(0.0, -0.5)$)--($(LongHorizonDecisionMaking)+(0.0, 0.5)$);
    
    \node[message, below left=0.5cm and -1.5cm of LongHorizonDecisionMaking] (Robot1DesiredTrajectory) {\shortstack{Desired Trajectory \\ (Long Horizon Plan)}};
    \node[decisionmakingmodule, below=0.5cm of Robot1DesiredTrajectory] (Robot1MediumHorizonDecisionMaking) {\shortstack{Medium Horizon Navigation \\ Decision Making (MH-NDM)}};
    \node[message, below=0.5cm of Robot1MediumHorizonDecisionMaking] (Robot1ReferenceTrajectory) {\shortstack{Reference Trajectory \\ (Medium Horizon Plan)}};
    \node[module, below=0.5cm of Robot1ReferenceTrajectory] (Robot1Controller) {Single-Robot Controller};
    \node[message, below=0.5cm of Robot1Controller] (Robot1DesiredControlInput) {\shortstack{Desired Control Input}};
    \node[decisionmakingmodule, below=0.5cm of Robot1DesiredControlInput] (Robot1ShortHorizonDecisionMaking) {\shortstack{Short Horizon Navigation \\ Decision Making (SH-NDM)}};
    \node[message, below=0.5cm of Robot1ShortHorizonDecisionMaking] (Robot1ControlInput) {\shortstack{Control Input}};
    \node[hardware, below=0.5cm of Robot1ControlInput] (Robot1Motors) {\shortstack{Actuators}};
    \node[message, left=of Robot1ReferenceTrajectory] (Robot1StaticObstacles) {\shortstack{Static Obstacles}};
    \node[message, below=0.5cm of Robot1StaticObstacles] (Robot1DynamicObstacles) {\shortstack{Dynamic Obstacles}};
    \node[message, below=0.5cm of Robot1DynamicObstacles] (Robot1CurrentEgoState) {\shortstack{Current Ego State}};
    \node[message, below=0.5cm of Robot1CurrentEgoState] (Robot1Teammates) {\shortstack{Teammates}};
    \node[abstract, above left=-0.5cm and 1.0cm of Robot1CurrentEgoState] (Robot1Sensing) {\shortstack{Perception \\ Localization \\ Mapping \\ Prediction}};
    \node[sensorshardware, above=0.5cm of Robot1Sensing] (Robot1Sensors) {\shortstack{Sensors}};
    \node[fit=(Robot1MediumHorizonDecisionMaking) (Robot1Motors) (Robot1Sensors), draw, inner sep=2mm] (Robot1Onboard) {};
    \node[normaltext, below left=-0.5cm and 6.5cm of Robot1Motors] (Robot1OnBoardText) {Robot 1 On-board};

    \node[message, below right=0.5cm and -1.5cm of LongHorizonDecisionMaking] (Robot2DesiredTrajectory) {\shortstack{Desired Trajectory \\ (Long Horizon Plan)}};
    \node[decisionmakingmodule, below=0.5cm of Robot2DesiredTrajectory] (Robot2MediumHorizonDecisionMaking) {\shortstack{Medium Horizon Navigation \\ Decision Making (MH-NDM)}};
    \node[message, below=0.5cm of Robot2MediumHorizonDecisionMaking] (Robot2ReferenceTrajectory) {\shortstack{Reference Trajectory \\ (Medium Horizon Plan)}};
    \node[module, below=0.5cm of Robot2ReferenceTrajectory] (Robot2Controller) {Single-Robot Controller};
    \node[message, below=0.5cm of Robot2Controller] (Robot2DesiredControlInput) {\shortstack{Desired Control Input}};
    \node[decisionmakingmodule, below=0.5cm of Robot2DesiredControlInput] (Robot2ShortHorizonDecisionMaking) {\shortstack{Short Horizon Navigation \\ Decision Making (SH-NDM)}};
    \node[message, below=0.5cm of Robot2ShortHorizonDecisionMaking] (Robot2ControlInput) {\shortstack{Control Input}};
    \node[hardware, below=0.5cm of Robot2ControlInput] (Robot2Motors) {\shortstack{Actuators}};
    \node[message, right=of Robot2ReferenceTrajectory] (Robot2StaticObstacles) {\shortstack{Static Obstacles}};
    \node[message, below=0.5cm of Robot2StaticObstacles] (Robot2DynamicObstacles) {\shortstack{Dynamic Obstacles}};
    \node[message, below=0.5cm of Robot2DynamicObstacles] (Robot2CurrentEgoState) {\shortstack{Current Ego State}};
    \node[message, below=0.5cm of Robot2CurrentEgoState] (Robot2Teammates) {\shortstack{Teammates}};
    \node[abstract, above right=-0.5cm and 1.0cm of Robot2CurrentEgoState] (Robot2Sensing) {\shortstack{Perception \\ Localization \\ Mapping \\ Prediction}};
    \node[sensorshardware, above=0.5cm of Robot2Sensing] (Robot2Sensors) {\shortstack{Sensors}};
    \node[fit=(Robot2MediumHorizonDecisionMaking) (Robot2Motors) (Robot2Sensors), draw, inner sep=2mm] (Robot2Onboard) {};
    \node[normaltext, below right=-0.5cm and 6.5cm of Robot2Motors] (Robot2OnBoardText) {Robot N On-board};

    \draw[->]  (LongHorizonDecisionMaking)--(Robot1DesiredTrajectory);
    \draw[dashed,->]  (Robot1DesiredTrajectory)--(Robot1MediumHorizonDecisionMaking);
    \draw[->]  (Robot1MediumHorizonDecisionMaking)--(Robot1ReferenceTrajectory);
    \draw[->]  (Robot1ReferenceTrajectory)--(Robot1Controller);
    \draw[->]  (Robot1Controller)--(Robot1DesiredControlInput);
    \draw[->]  (Robot1DesiredControlInput)--(Robot1ShortHorizonDecisionMaking);
    \draw[->]  (Robot1ShortHorizonDecisionMaking)--(Robot1ControlInput);
    \draw[->]  (Robot1ControlInput)--(Robot1Motors);
    \draw[->]  (Robot1StaticObstacles)--($(Robot1StaticObstacles)+(2.5, 0.0)$)--($(Robot1StaticObstacles)+(3.5, 1.5)$);
    \draw[->]  (Robot1DynamicObstacles)--($(Robot1DynamicObstacles)+(2.5, 0.0)$)--($(Robot1DynamicObstacles)+(3.5, 3.0)$);
    \draw[->]  (Robot1CurrentEgoState)--($(Robot1CurrentEgoState)+(2.5, 0.0)$)--($(Robot1CurrentEgoState)+(3.5, 4.5)$);
    \draw[->]  (Robot1CurrentEgoState)--($(Robot1CurrentEgoState)+(2.5, 0.0)$)--($(Robot1CurrentEgoState)+(3.5, 1.5)$);
    \draw[->]  (Robot1Teammates)--($(Robot1Teammates)+(2.5, 0.0)$)--($(Robot1Teammates)+(3.5, 6.0)$);
    \draw[->]  (Robot1StaticObstacles)--($(Robot1StaticObstacles)+(2.5, 0.0)$)--($(Robot1StaticObstacles)+(3.5, -4.5)$);
    \draw[->]  (Robot1DynamicObstacles)--($(Robot1DynamicObstacles)+(2.5, 0.0)$)--($(Robot1DynamicObstacles)+(3.5, -3.0)$);
    \draw[->]  (Robot1CurrentEgoState)--($(Robot1CurrentEgoState)+(2.5, 0.0)$)--($(Robot1CurrentEgoState)+(3.5, -1.5)$);
    \draw[->]  (Robot1Teammates)--($(Robot1Teammates)+(2.5, 0.0)$)--($(Robot1Teammates)+(3.5, 0.0)$);
    \draw[->] (Robot1Sensors)--(Robot1Sensing);
    \draw[->] (Robot1Sensing)--($(Robot1Sensing)+(1.3, 0.0)$)--($(Robot1Sensing)+(2.3, 2.0)$);
    \draw[->] (Robot1Sensing)--($(Robot1Sensing)+(1.3, 0.0)$)--($(Robot1Sensing)+(2.3, 0.5)$);
    \draw[->] (Robot1Sensing)--($(Robot1Sensing)+(1.3, 0.0)$)--($(Robot1Sensing)+(2.3, -1.0)$);
    \draw[->] (Robot1Sensing)--($(Robot1Sensing)+(1.3, 0.0)$)--($(Robot1Sensing)+(2.3, -2.5)$);
    \draw[->] (Robot1ControlInput)--($(Robot1Sensing)+(0.0, -4.0)$)--(Robot1Sensing);
    \draw[dashed,->] (Robot1StaticObstacles)--($(Robot1StaticObstacles)+(0.0, 4.5)$)--(LongHorizonDecisionMaking);
    \draw[dashed,->]  (Robot1CurrentEgoState)--($(Robot1CurrentEgoState)+(0.0, -0.75)$)--($(Robot1CurrentEgoState)+(-6.5, -0.75)$)--($(Robot1CurrentEgoState)+(-6.5, 7.5)$)--(LongHorizonDecisionMaking);
    
    \draw[->]  (LongHorizonDecisionMaking)--(Robot2DesiredTrajectory);
    \draw[dashed,->]  (Robot2DesiredTrajectory)--(Robot2MediumHorizonDecisionMaking);
    \draw[->]  (Robot2MediumHorizonDecisionMaking)--(Robot2ReferenceTrajectory);
    \draw[->]  (Robot2ReferenceTrajectory)--(Robot2Controller);
    \draw[->]  (Robot2Controller)--(Robot2DesiredControlInput);
    \draw[->]  (Robot2DesiredControlInput)--(Robot2ShortHorizonDecisionMaking);
    \draw[->]  (Robot2ShortHorizonDecisionMaking)--(Robot2ControlInput);
    \draw[->]  (Robot2ControlInput)--(Robot2Motors);
    \draw[->]  (Robot2StaticObstacles)--($(Robot2StaticObstacles)+(-2.5, 0.0)$)--($(Robot2StaticObstacles)+(-3.5, 1.5)$);
    \draw[->]  (Robot2DynamicObstacles)--($(Robot2DynamicObstacles)+(-2.5, 0.0)$)--($(Robot2DynamicObstacles)+(-3.5, 3.0)$);
    \draw[->]  (Robot2CurrentEgoState)--($(Robot2CurrentEgoState)+(-2.5, 0.0)$)--($(Robot2CurrentEgoState)+(-3.5, 4.5)$);
    \draw[->]  (Robot2CurrentEgoState)--($(Robot2CurrentEgoState)+(-2.5, 0.0)$)--($(Robot2CurrentEgoState)+(-3.5, 1.5)$);
    \draw[->]  (Robot2Teammates)--($(Robot2Teammates)+(-2.5, 0.0)$)--($(Robot2Teammates)+(-3.5, 6.0)$);
    \draw[->]  (Robot2StaticObstacles)--($(Robot2StaticObstacles)+(-2.5, 0.0)$)--($(Robot2StaticObstacles)+(-3.5, -4.5)$);
    \draw[->]  (Robot2DynamicObstacles)--($(Robot2DynamicObstacles)+(-2.5, 0.0)$)--($(Robot2DynamicObstacles)+(-3.5, -3.0)$);
    \draw[->]  (Robot2CurrentEgoState)--($(Robot2CurrentEgoState)+(-2.5, 0.0)$)--($(Robot2CurrentEgoState)+(-3.5, -1.5)$);
    \draw[->]  (Robot2Teammates)--($(Robot2Teammates)+(-2.5, 0.0)$)--($(Robot2Teammates)+(-3.5, 0.0)$);
    \draw[->] (Robot2Sensors)--(Robot2Sensing);
    \draw[->] (Robot2Sensing)--($(Robot2Sensing)+(-1.3, 0.0)$)--($(Robot2Sensing)+(-2.3, 2.0)$);
    \draw[->] (Robot2Sensing)--($(Robot2Sensing)+(-1.3, 0.0)$)--($(Robot2Sensing)+(-2.3, 0.5)$);
    \draw[->] (Robot2Sensing)--($(Robot2Sensing)+(-1.3, 0.0)$)--($(Robot2Sensing)+(-2.3, -1.0)$);
    \draw[->] (Robot2Sensing)--($(Robot2Sensing)+(-1.3, 0.0)$)--($(Robot2Sensing)+(-2.3, -2.5)$);
    \draw[->] (Robot2ControlInput)--($(Robot2Sensing)+(0.0, -4.0)$)--(Robot2Sensing);    \draw[dashed,->] (Robot2StaticObstacles)--($(Robot2StaticObstacles)+(0.0, 4.5)$)--(LongHorizonDecisionMaking);
    \draw[dashed,->]  (Robot2CurrentEgoState)--($(Robot2CurrentEgoState)+(0.0, -0.75)$)--($(Robot2CurrentEgoState)+(6.5, -0.75)$)--($(Robot2CurrentEgoState)+(6.5, 7.5)$)--(LongHorizonDecisionMaking);

    \draw[<->, decorate, decoration=snake] (Robot1MediumHorizonDecisionMaking)--(Robot2MediumHorizonDecisionMaking);
    \draw[<->, decorate, decoration=snake] (Robot1ShortHorizonDecisionMaking)--(Robot2ShortHorizonDecisionMaking);




\end{tikzpicture}}
\vspace{-0.25in}
\caption{\textbf{System Design.} Our system utilizes long, medium, and short horizon decision making modules to allow collision and deadlock-free operation. The {\color{violet}purple} boxes are the messages passed between components, the {\color{brown} brown} boxes are decision making modules, and the {\color{teal}teal} box is the controller. The physical robot parts are denoted with {\color{darkgray}gray} boxes, and the abstract perception, localization, mapping, and prediction subsystems are shown as a single white box. Solid lines represent connections on the same machine, dashed lines represent connections over a communication medium, and cannot be expected to be always possible, and lastly, squiggly lines represent interactions between components either explicitly through communication, or implicitly using the fact that robots use the same algorithm. The long horizon decision making module is centrally deployed, and other components run on-board.\looseness=-1}
\label{Figure:SystemDesign}
\vspace{-0.25in}
\end{figure*}
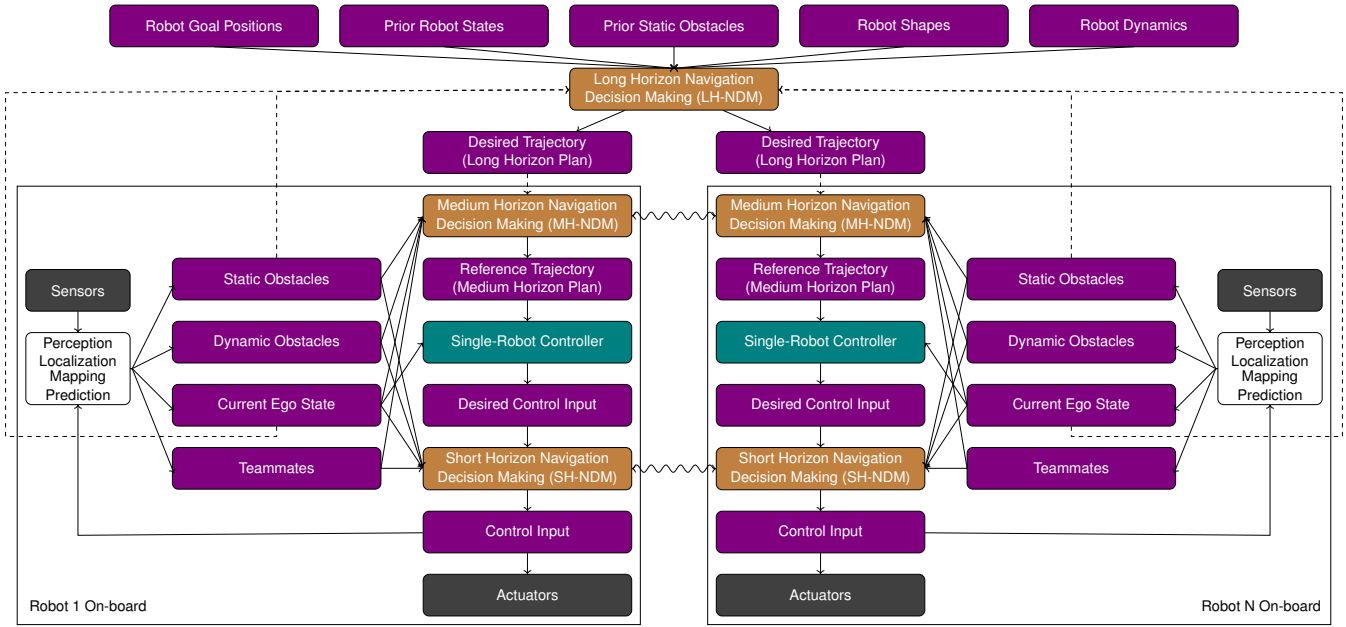

We describe the abstract system design of MRNAV (\Cref{Figure:SystemDesign}), its components, responsibilities assigned to the components, and their deployment options.\looseness=-1

\textbf{Perception, Localization, Mapping, and Prediction Interfaces.} The robots' on-board perception, localization, mapping, and prediction systems consume sensor data and the control inputs applied to the actuators to i) build representations for static and dynamic obstacles and teammates, and ii) estimate the robot states.
Generally, static obstacles are modeled using their positions and shapes, dynamic obstacles are modeled using their states, shapes, and possibly future behavior predictions, and teammates are modeled using their states, shapes, and possibly estimated or communicated future trajectories.\looseness=-1

\textbf{Long Horizon Navigation Decision Making.} An LH-NDM module is deployed on a centralized computing system, which is responsible for generating \emph{desired trajectories}.
It sends the computed desired trajectories to the robots through a communication link, which the robots follow as closely as possible using their on-board decision making system, while ensuring that they remain collision-free, deviating from the desired trajectories as necessary, but navigating back to them whenever possible.
This allows robots to delegate global navigation decisions to the LH-NDM module, and locally resolve collisions not accounted for by the desired trajectories in real-time.
If reliable communication to the central entity is possible, robots send their local static obstacle representations to the centralized system, in order for it to update its global environment representation.
Also, the robots send their estimated states to the centralized system, to allow it to track task progress and re-plan a new desired trajectory if i) a previous navigation task is completed, and/or ii) a deadlock cannot be resolved locally.
The LH-NDM module can be given robot goal positions explicitly if the navigation system is used by an outside task planner, or it can compute goal positions itself if it does task planning as well.
Desired trajectories can be made dynamically feasible if desired.
The long horizon decision making module runs on-demand, and does not have any real-time requirements, allowing it to provide completeness and optimality guarantees if desired.\looseness=-1

\textbf{Medium Horizon Navigation Decision Making.} Robots employ on-board MH-NDM modules for medium term collision avoidance and local deadlock resolution.
These modules generate \emph{reference trajectories} in real-time for robots' controllers' to track.
Dynamic feasibility of the generated reference trajectories is desirable so that the downstream single-robot controller can track them without major trajectory tracking imperfections, which allows the system to realize the collision avoidance and deadlock resolution behavior of the MH-NDM.
This module uses the desired trajectories as guidance, such that it generates reference trajectories that are close to the desired trajectories, but allows deviations from them as necessary.
Since the robot can maintain only the local representation of the environment because of memory and storage limitations, and can reason about a limited horizon during decision making because of the real-time operation, the reference trajectories are computed in a receding horizon fashion.
The MH-NDM modules across multiple robots can interact with each other either explicitly through communication, and/or implicitly using the fact that all teammates run the same algorithm to allow inter-robot collision avoidance and cooperation.
The module should not depend on perfect instantaneous communication assumption, and be robust to communication imperfections.\looseness=-1

\textbf{Short Horizon Navigation Decision Making.} The single-robot controller computes control actions to track the reference trajectories. 
However, collision avoidance behavior enforced by MH-NDM module is not enough for safe operation.
The reasons are i) medium horizon planners ensure safe operation not for the full navigation task but for the medium term future and all state-of-the-art medium horizon planners fail occasionally because the solved problem has no feasibility guarantee, which may cause reference trajectories to be unsafe when they are used for a long duration, and ii) even if the reference trajectories are dynamically feasible, robots may diverge from them because of state estimation inaccuracies, unmodeled dynamics, disturbances, or imperfect system identification. 
In order to allow collision-free operation in these cases, we utilize the SH-NDM module.
This module is a high frequency safety module that takes control inputs generated by the controller, which we call the \emph{desired control inputs}, and makes them safe in a minimally invasive manner, computing safe control inputs that are as close as possible to the desired control inputs, which are subsequently sent to the actuators.
SH-NDM modules do not reason about deadlock-free operation as that responsibility is shared between the LH-NDM and MH-NDM modules.
Its main aim is to ensure that the control inputs sent to the actuators do not result in collisions.
Similarly to the MH-NDM modules, the SH-NDM modules across multiple robots can interact with each other explicitly or implicitly, and they should not depend on perfect communication assumption to enforce safety.\looseness=-1

\vspace{0.05in}

The system (\Cref{Figure:SystemDesign}) is designed in such a way that desired trajectories computed by the LH-NDM guide the MH-NDM modules, and the desired control inputs generated to track the reference trajectories guide the SH-NDM module.
The inputs and outputs of on-board safety modules have the same structure, which allows removing MH-NDM or SH-NDM modules by connecting their inputs to their outputs, if the benefits they provide are not needed.
In general, i) LH-NDM module should aim to provide desired trajectories that results in low spatiotemporal deviation of the reference trajectories computed by the MH-NDM from the desired ones, and ii) the MH-NDM module should aim to provide reference trajectories that results in low spatiotemporal deviation of the robot position from the reference trajectories after executing the actions generated by the SH-NDM module in order for effective navigation.
Otherwise, robots will take a considerably longer time to reach their goal positions, as the lower level decision making modules will override higher level decisions in order to enforce safety.\looseness=-1

\subsection{Implementation for Quadrotor Flight}\label{Section:ImplementationForQuadrotorFlight}


We implement our design for quadrotor flight in simulations using the RotorS simulator~\cite{furrer2016rotors} and the Robot Operating System (ROS) in C++.
We simulate physics using the Open Dynamics Engine (ODE)\footnote{\url{http://ode.org/}}.
The quadrotors are controlled via speed commands for each of the four motors.
We release the code of our system\footnote{The code will be released upon acceptance.}, which contains implementation of the navigation stack and algorithms used within.\looseness=-1

We use octrees~\cite{hornung2013octomap} to represent static obstacles.
We use position and velocities as the sensed states of the teammates.
Dynamic obstacle positions, velocities, and shapes are sensed by the teammates.
Each object's shape is modeled as an axis aligned box.\looseness=-1

\textbf{Long Horizon Navigation Decision Making.}
The LH-NDM module randomly generates goal positions for the robots, computes shortest paths from robots' current positions to the goal positions avoiding only the prior static obstacles, which we set to either the actual map of the environment or an empty map in the experiments, using A* search and assigns durations to each segment of the computed path.
Our LH-NDM module does not enforce dynamic feasibility, because LH-NDM module does not enforce collision avoidance against dynamic obstacles or teammates, which requires necessary divergence from desired trajectories during execution.
When a robot reaches to its goal position, the LH-NDM module randomly generates a new goal position, and computes a new desired trajectory.\looseness=-1

\textbf{Medium Horizon Navigation Decision Making.}
We use DREAM~\cite{senbaslar2023dream} as the MH-NDM module.
DREAM is a \underline{d}ecentralized \underline{re}al-time \underline{a}synchronous trajectory planning algorithm for \underline{m}ulti-robot teams, and computes reference trajectories that avoids static obstacles, dynamic obstacles, as well as teammates.
For static obstacle avoidance, it requires shapes of the obstacles and their existence probabilities.\footnote{We provide the static obstacles from the octree. We do not implement probabilistic sensing, hence, static obstacles have existence probability $1.0$.\looseness=-1}.\looseness=-1

For dynamic obstacle avoidance, DREAM requires i) current shapes and positions of the dynamic obstacles, and ii) a probability distribution across possible behavior models.
A behavior model $\mB = (\mM, \mI)$ is a tuple of movement model $\mM$ and interaction model $\mI$.
A movement model $\mM: \mathbb{R}^3 \rightarrow \mathbb{R}^3$ is a function mapping the dynamic obstacle's position to its desired velocity, describing the intention of the obstacle.
An interaction model $\mI: \mathbb{R}^{12}\rightarrow \mathbb{R}^3$ is a function describing the interaction of the dynamic obstacle with teammates, mapping the dynamic obstacle's position, desired velocity, the ego robot's position and velocity to the obstacle's velocity.
In the simulations, each dynamic obstacle is modeled as running its movement model to obtain its velocity\footnote{In this paper, we do not use interactive dynamic obstacles in order to put collision avoidance responsibility to the planning robots.\looseness=-1}.
The on-board systems do not have explicit access to the behavior models of the dynamic obstacles, and use on-board probabilistic behavior model estimation instead~\cite{senbaslar2023dream}.\looseness=-1

For teammate avoidance, DREAM relies on discretized separating hyperplane trajectories (DSHTs)~\cite{senbaslar2022async}, which are the trajectories of separating hyperplanes between two moving bodies, discretized at a high frequency. 
DSHT based constraints allow the active trajectories of each pair of robots to share a separating hyperplane constraint at all times under asynchronous planning and communication imperfections~\cite{senbaslar2022async}.
Robots prune DSHTs whenever they know another robot has planned successfully.
Whenever MH-NDM plans successfully, it broadcasts a message to other robots.
If others receive this message, they prune the DSHT between the sender robot and themselves as explained in~\cite{senbaslar2022async,senbaslar2023dream}.\looseness=-1

DREAM employs a widely used goal selection, discrete search, and trajectory optimization pipeline.
In each planning iteration, the goal selection step chooses a goal position on the desired trajectory to plan to.
In the discrete search stage, DREAM computes a spatiotemporal path to the selected goal position by minimizing collision probabilities with static obstacles, dynamic obstacles, as well as DSHT violations.
In the trajectory optimization stage, quadratic programming based spline optimization is executed, making the computed discrete spatiotemporal path dynamically feasible while preserving collision probabilities computed and DSHT hyperplanes not violated during search.\looseness=-1

\textbf{Controller and Short Horizon Decision Making.}
The RotorS simulator accepts angular motor speed inputs for each of the four propellers of the quadrotor.
The forces and moments applied by the propellers are simulated based on the model of a single propeller near hovering~\cite{martin2010acc}.
Our design (Fig.~\ref{Figure:SystemDesign}) necessitates a SH-NDM module that accepts desired control inputs and outputs safe control inputs in a minimally invasive manner, considering static and dynamic obstacles as well as teammates.
We integrate a quadrotor controller~\cite{lee2010control} that internally computes desired acceleration commands with an safety barrier certificates (SBC) based SH-NDM module for double integrator systems~\cite{wang2017safety} allowing static and dynamic obstacle and cooperative teammate avoidance\footnote{Some alternatives are~\cite{li2017flatctrl}, which makes the reference trajectories safe utilizing SBCs, and~\cite{wu2016safety}, which computes forces and moments directly, but does not explicitly model multiple robots. We use~\cite{wang2017safety} because it explicitly accounts for multiple robots and has extensive evaluations\looseness=-1}.
By doing so, we diverge from the proposed design slightly, because we do not ensure the safety of the final motor speeds, but ensure that the intermediate accelerations computed by the controllers are safe.\looseness=-1

\begin{figure}[t]
    \centering
    \includegraphics[width=\linewidth]{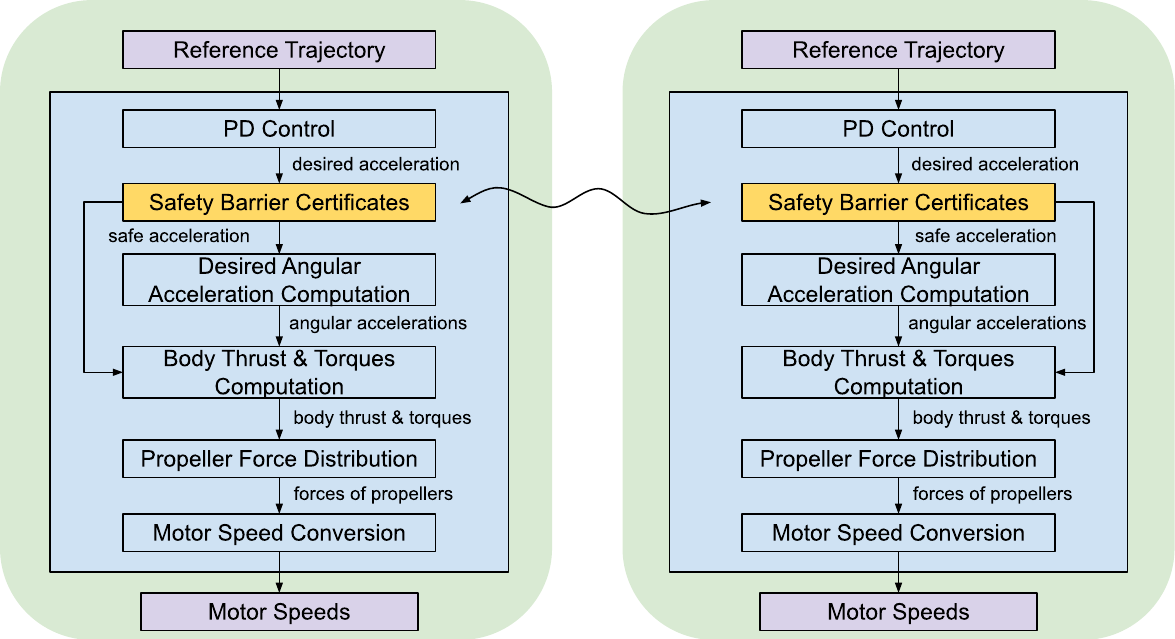}
    \vspace{-0.25in}
    \caption{\textbf{Controller and SH-NDM Module Integration.} Safety barrier certificates based SH-NDM module~\cite{wang2017safety} assumes double integrator dynamics, i.e., that robots can be controlled with acceleration commands. Since quadrotors cannot be directly controlled with acceleration commands, we augment a quadrotor controller~\cite{lee2010control} with SBC by inserting it after desired acceleration computation within the controller. The SH-NDM modules in different robots interact assuming that all robots run the same algorithm.\looseness=-1}
    \label{Figure:SafeController}
    \vspace{-0.25in}
\end{figure}

The controller computes a desired body acceleration using a PD controller to track reference trajectory samples composed of desired position, velocity, and yaw angles.
We set desired yaw angles to zero.
Then, the controller computes the desired angular acceleration of the quadrotor in each of the three frame axes in order to produce the given desired acceleration and yaw.
Afterwards, it computes the desired total body thrust and torques to achieve the desired acceleration and angular accelerations.
Next, it computes the forces that need to be generated by the propellers in order to produce the desired total body thrust and torques. 
The conversion of the desired forces to motor speed commands are done similarly to the simulator.
We insert the SH-NDM module~\cite{wang2017safety} within the controller after desired body acceleration computation (\Cref{Figure:SafeController}).\looseness=-1

\section{Evaluation}

In order to show the efficacy of our design, i) we ablate long, medium, and short horizon decision making modules and show that all components are required for collision and deadlock-free navigation, and ii) we run our system in an uninterrupted eight hours wall time (six hours simulation time) long simulation with static and dynamic obstacles as well as multiple robots without collisions and deadlocks.\looseness=-1

\subsection{Ablation Study}

In our ablation study, eight quadrotors navigate in a forest-like or a maze-like environment.
The maze-like environment contains concave regions, in which the robots may stall and deadlock.
Dynamic obstacles are randomly generated.
Each dynamic obstacle has an axis aligned box shape with a diagonal length sampled uniformly in $[\SI{1.2}{m}, \SI{1.6}{m}]$.
Movement models generate desired velocities so that the dynamic obstacles rotate around the line going perpendicular to the x-y plane and goes through the origin.
The norm of the desired velocities are sampled uniformly in $[\SI{0.25}{\frac{m}{s}}, \SI{0.35}{\frac{m}{s}}]$ once, i.e., they move with constant speed.\looseness=-1

We run each experiment for $\SI{300}{s}$ in simulation time.
Each randomly generated goal position by the LH-NDM module is called a task that should be navigated to.
When we ablate the SH-NDM module, we set safe acceleration to the desired acceleration (Fig.~\ref{Figure:SafeController}).
When we ablate the MH-NDM module, we set reference trajectory to the desired trajectory.
When we ablate the LH-NDM module, we provide an empty map as the prior static obstacles, which causes LH-NDM to produce straight lines to the goal positions as desired trajectories, otherwise we provide the real map to it.\looseness=-1

We use four evaluation metrics: i) deadlock rate (\textbf{Deadl. R.}), the ratio of robots that deadlock at the end of the experiment, ii) collision rate (\textbf{Coll. R.}), the ratio of tasks during which the robot has collided with an object, iii) completion rate (\textbf{Comp. R.}), the ratio of tasks robots complete without a collision, and iv) average navigation duration (\textbf{Avg. N. Dur.}), the average duration for robots to complete no-collision tasks\footnote{Deadlock, collision, and completion rates are primary metrics to compare module combinations. The average navigation duration is a secondary metric and is only meaningful when there are no deadlocks or collisions.\looseness=-1}.\looseness=-1

\begin{figure}
    \centering
     \subfloat[Forest with 400 Dyn. Obstacles]{%
       \includegraphics[width=0.45\linewidth]{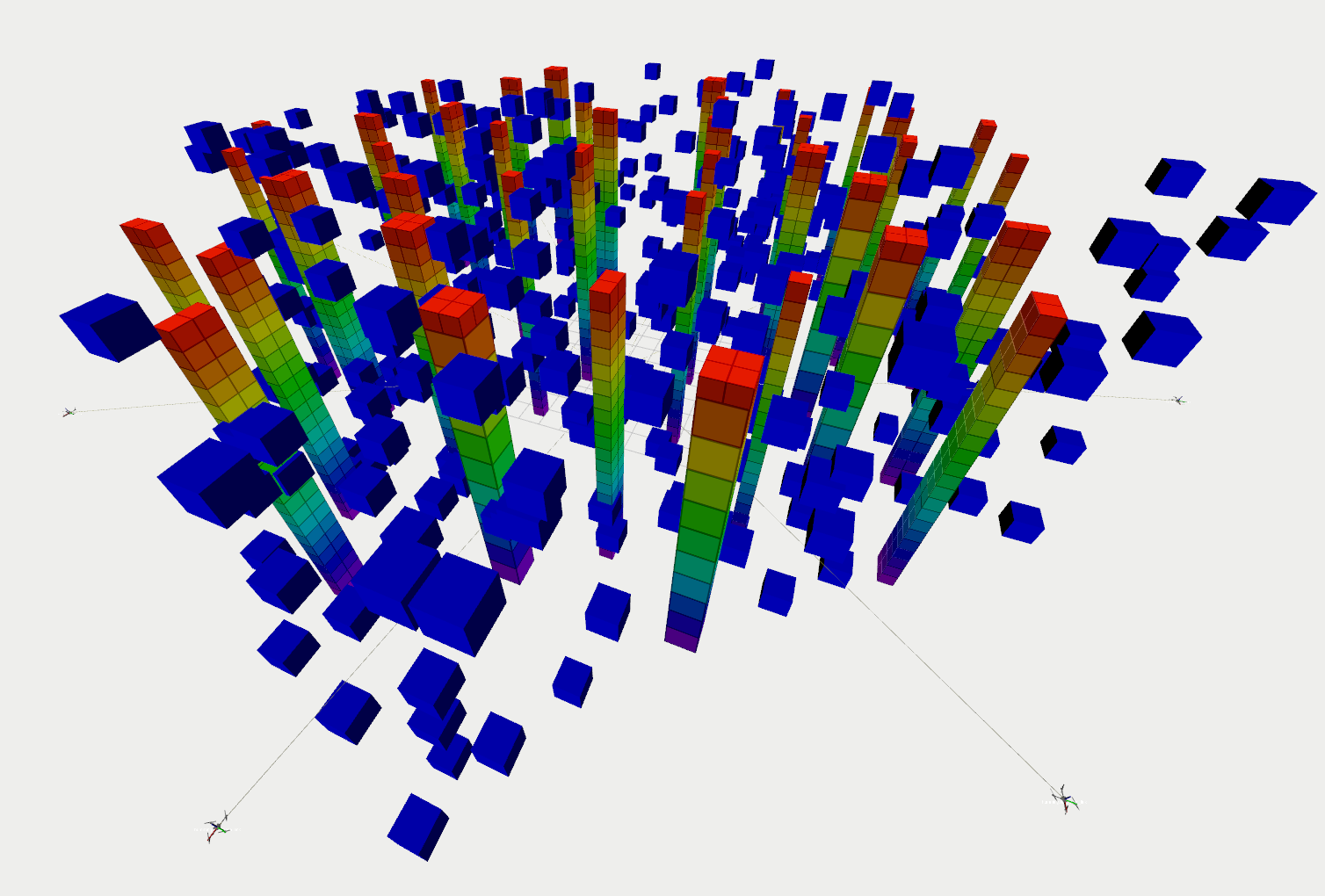}
       \label{Figure:AblationStudyForest400}
     }
     \hfill
     \subfloat[Forest with No Dyn. Obstacles]{%
       \includegraphics[width=0.45\linewidth]{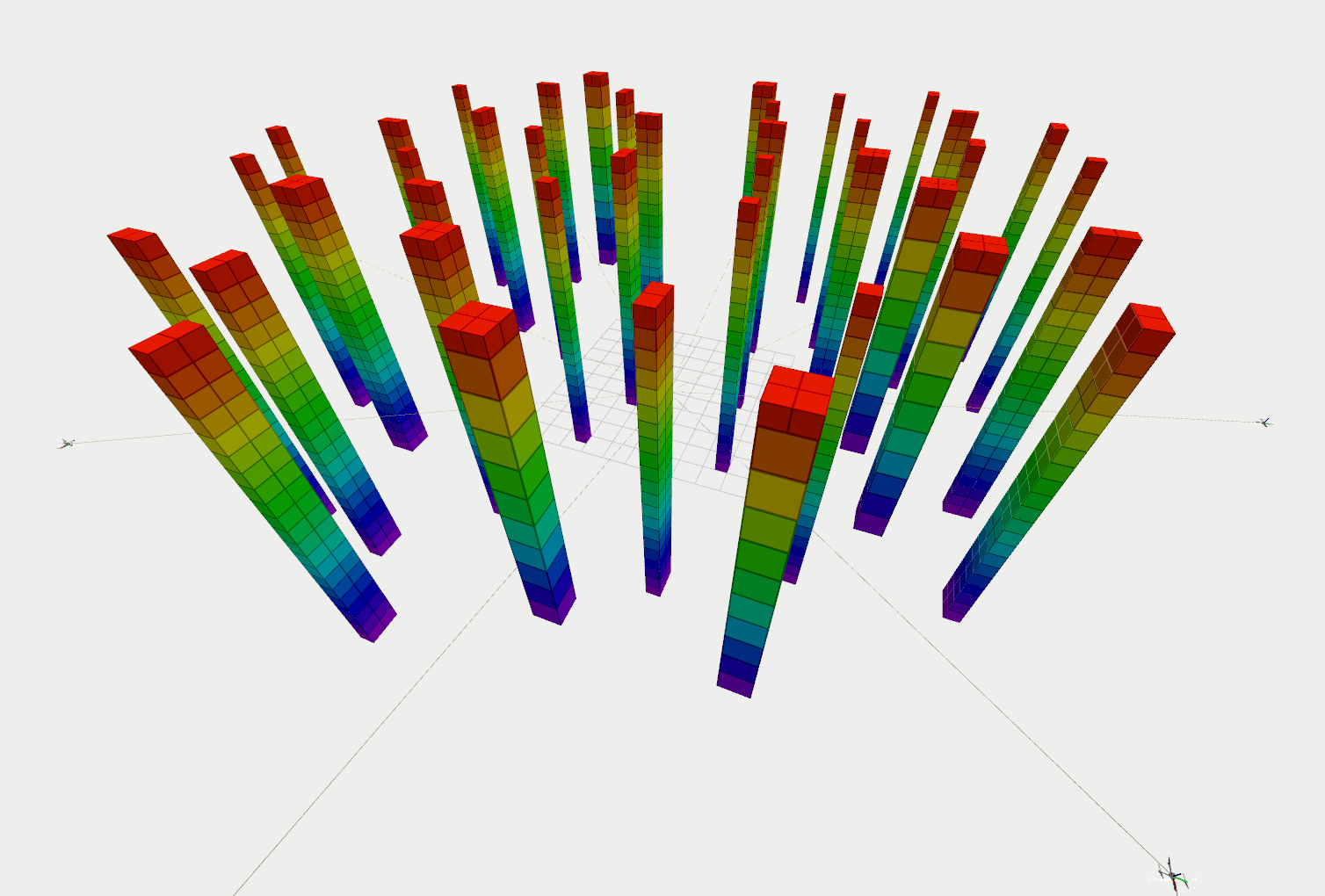}
       \label{Figure:AblationStudyForest0}
     }
     \\
     \vspace{-0.1in}
    \subfloat[Maze with 200 Dyn. Obstacles]{%
       \includegraphics[width=0.45\linewidth]{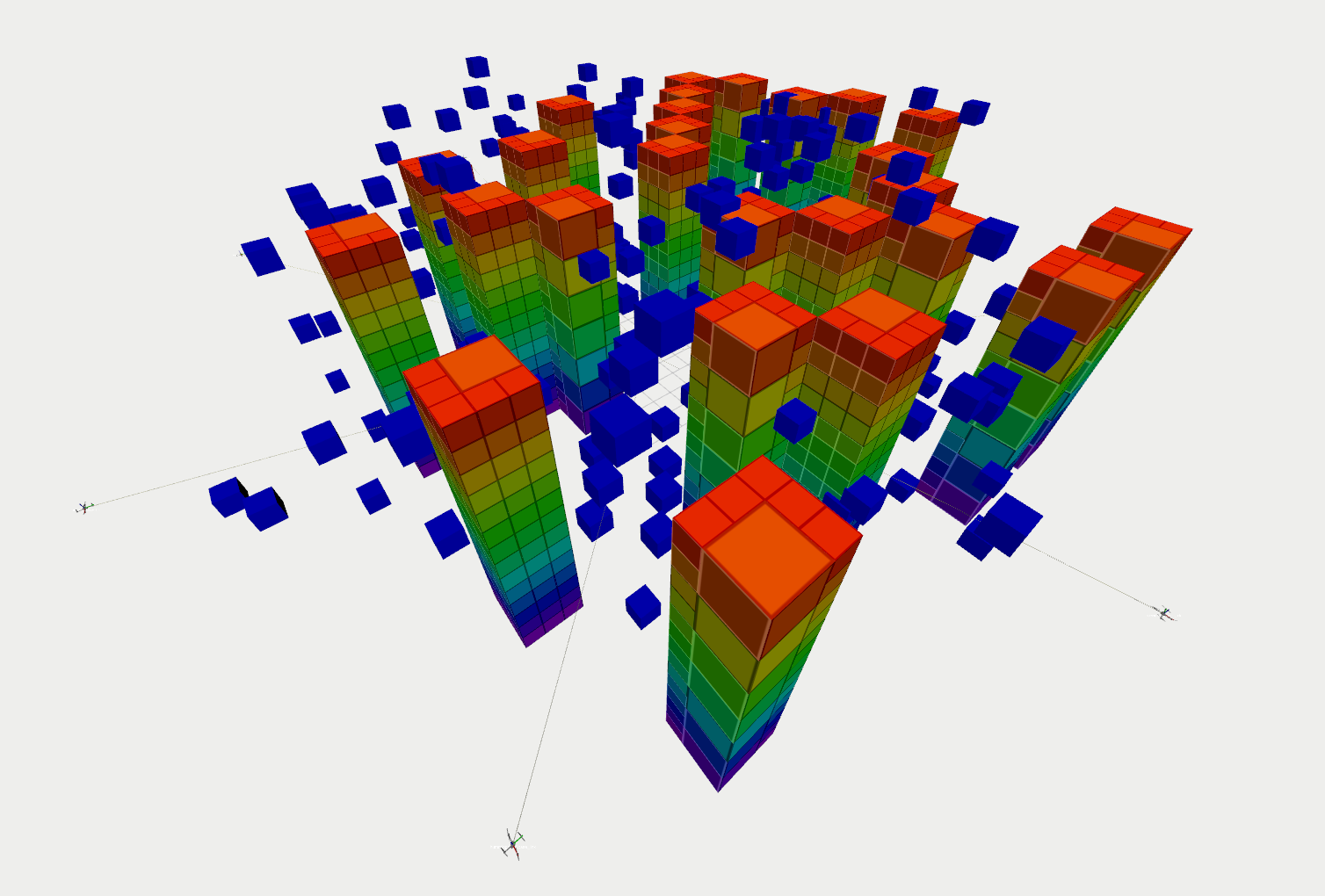}
       \label{Figure:AblationStudyMaze200}
     }
     \hfill
     \subfloat[Maze with No Dyn. Obstacles]{%
       \includegraphics[width=0.45\linewidth]{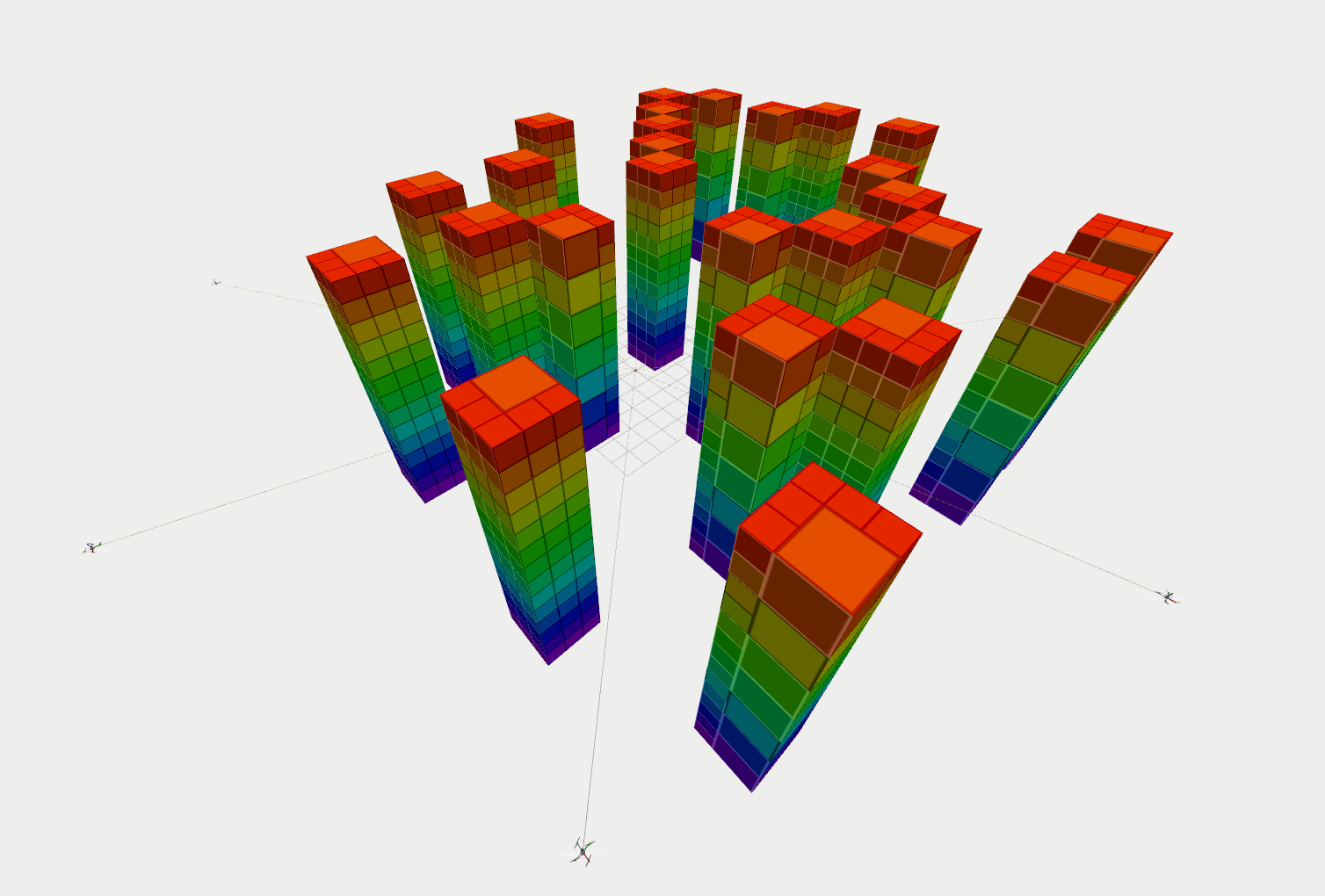}
       \label{Figure:AblationStudyMaze0}
     }
     \caption{The environments used in the ablation study. The dynamic obstacles are {\color{blue}blue} boxes, and static obstacles are the pillars.}
     \label{Figure:AblationStudy}
     \vspace{-0.05in}
\end{figure}

\begin{table}[t]
    \centering
    \caption{Forest environment with $400$ dynamic obstacles}
    \vspace{-0.1in}
    \resizebox{\linewidth}{!}{\begin{tabular}{c|c|c|c|c}
         Modules & Deadl. R.& Coll. R. & Comp. R. &  Avg. N. Dur. [s] \\
         \hline SH & $0/8$ & $5/132$ & $127/132$ & $13.19$\\
         MH & $0/8$ & $73/216$ & $143/216$ & $10.52$\\
         LH & $0/8$ & $246/313$ & $67/246$ & $4.34$ \\
         \textbf{MH+SH} & $0/8$ & $0/148$ & $148/148$ & $15.45$\\
         LH+SH & $2/8$ & $4/159$ & $155/159$ & $11.93$\\
         LH+MH & $0/8$ & $102/220$ & $118/220$ & $10.13$\\
         \textbf{LH+MH+SH} & $0/8$ & $0/158$ & $158/158$ & $14.79$
    \end{tabular}}
    \label{Table:Forest400Dyn8Quad}
    \vspace{-0.25in}
\end{table}

\begin{table}[t]
    \centering
    \caption{Forest environment with no dynamic obstacles}
    \vspace{-0.11in}
    \resizebox{\linewidth}{!}{\begin{tabular}{c|c|c|c|c}
         Modules & Deadl. R.& Coll. R. & Comp. R. &  Avg. N. Dur. [s] \\
         \hline SH & $6/8$ & $0/158$ & $158/158$ & $11.43$ \\
         MH & $0/8$ & $4/167$ & $163/167$ &  $13.93$\\
         LH & $0/8$ & $37/313$ & $276/313$ & $7.30$\\
         \textbf{MH+SH} & $0/8$ & $0/157$ & $157/157$ & $13.04$\\
         \textbf{LH+SH} & $0/8$ & $0/335$ & $335/335$ & $6.99$\\
         LH+MH & $0/8$ & $3/193$ & $190/193$ & $11.94$\\
         \textbf{LH+MH+SH} & $0/8$ & $0/185$ & $185/185$ & $12.95$\\
    \end{tabular}}
    \label{Table:Forest0Dyn8Quad}
    \vspace{-0.26in}
\end{table}

\begin{table}[t]
    \centering
    \caption{Maze environment with $200$ dynamic obstacle}
    \vspace{-0.1in}
    \resizebox{\linewidth}{!}{\begin{tabular}{c|c|c|c|c}
         Modules & Deadl. R.& Coll. R. & Comp. R. &  Avg. N. Dur. [s] \\
         \hline SH & $8/8$ & $12/39$ & $27/39$ & $14.84$\\
         MH & $2/8$ & $49/107$ & $58/107$ & $12.27$\\
         LH & $0/8$ & $149/286$ & $137/286$ & $6.74$\\
         MH+SH & $7/8$ & $6/22$ & $16/22$ & $15.21$\\
         LH+SH & $6/8$ & $2/141$ & $139/141$ & $10.38$\\
         LH+MH & $0/8$ & $29/183$ & $154/183$ & $12.23$\\
         \textbf{LH+MH+SH} & $0/8$ & $0/142$ & $142/142$ & $16.20$
    \end{tabular}}
    \label{Table:Maze200Dyn8Quad}
    \vspace{-0.05in}
\end{table}

\begin{table}[t]
    \centering
    \caption{Maze environment with no dynamic obstacles}
    \vspace{-0.1in}
    \resizebox{\linewidth}{!}{\begin{tabular}{c|c|c|c|c}
         Modules & Deadl. R.& Coll. R. & Comp. R. &  Avg. N. Dur. [s] \\
         \hline SH & $8/8$ & $0/20$ & $20/20$ & $6.19$\\
         MH & $6/8$ & $16/42$ & $26/42$ & $15.54$\\
         LH & $0/8$ & $43/287$ & $244/287$ & $7.81$\\
         MH+SH & $8/8$ & $0/4$ & $4/4$ & $16.46$\\
         \textbf{LH+SH} & $0/8$ & $0/274$ & $274/274$ & $8.49$\\
         LH+MH & $0/8$ & $6/145$ & $139/145$ & $15.76$\\
         \textbf{LH+MH+SH} & $0/8$ & $0/151$ & $151/151$ & $15.44$
    \end{tabular}}
    \label{Table:Maze0Dyn8Quad}
    \vspace{-0.25in}
\end{table}

There are four cases we use: i) forest-like environment with $400$ dynamic obstacles (Fig.~\ref{Figure:AblationStudyForest400}), or ii) no dynamic obstacles (Fig.~\ref{Figure:AblationStudyForest0}), and iii) maze-like environment with $200$ dynamic obstacles (Fig.~\ref{Figure:AblationStudyMaze200}), or iv) no dynamic obstacles (Fig.~\ref{Figure:AblationStudyMaze0}).
The results of our experiments are summarized in~\Cref{Table:Forest400Dyn8Quad,Table:Forest0Dyn8Quad,Table:Maze200Dyn8Quad,Table:Maze0Dyn8Quad}.
Module combinations that result in zero deadlock and collision rates are highlighted in the tables.
Recordings can be found in the supplemental videos.\looseness=-1

\textbf{Enforcing safety of each executed action is essential.}
The SH-NDM module enforces that each action executed is safe.
We observe that this is an essential property for the effectiveness of the design.
All successful combinations contain the SH-NDM module.
It i) enables highly reactive collision avoidance behavior, ii) ensures safety when the MH-NDM fails, and iii) allows arbitrary divergence from the desired and reference trajectories without giving up safety.\looseness=-1

\textbf{Dynamic obstacles create the need for local deadlock resolution.}
When there are no dynamic obstacles, LH+SH module combination is successful in both forest and maze-like environments.
Since there are no dynamic obstacles, desired trajectories generated by the LH-NDM module avoid all obstacles.
SH-NDM module is enough to locally resolve collisions between teammates, after which, the robot is able to navigate back to its desired trajectory.\looseness=-1

When dynamic obstacles are added to the environment, LH+SH is not effective anymore because the desired trajectories do not avoid dynamic obstacles, and hence, dynamic obstacle avoidance responsibility is given to the SH-NDM module.
This causes robots to diverge from their desired trajectories considerably, causing robots to not be able to navigate back to them using only SH-NDM, resulting in deadlocks.
The LH+MH+SH combination is effective in such scenarios, because the MH-NDM allows local deadlock resolution.\looseness=-1

\textbf{Medium horizon decision making increases navigation duration.}
If the LH+SH module combination is successful, adding MH-NDM increases the average navigation duration (\Cref{Table:Forest0Dyn8Quad,Table:Maze0Dyn8Quad}), because, we impose lower dynamic limits on the reference trajectories than the actual dynamic limits of the quadrotors for the real-time executability during MH-NDM.
Otherwise, the planner needs to generate spatially longer trajectories with the same temporal horizon, which increases the run-time of the algorithm to achieve a similar collision avoidance performance.\looseness=-1

\textbf{Global deadlock resolution requires global reasoning.}
In the maze environment, all successful combinations include LH whether regardless of dynamic obstacles.
This is the case because the robot needs to reason about long trajectories to navigate in such environments.
The system needs to provide global reasoning capability for global deadlock resolution.\looseness=-1

\textbf{No approach is sufficient by itself and all is required for effectiveness in all cases.}
None of LH, MH, or SH-NDM modules are sufficient for navigating collision and deadlock-free in neither forest nor maze-like environments by themselves.
The LH+MH+SH combination is the only one that allows collision and deadlock-free navigation in all cases.\looseness=-1

\subsection{Long Term Execution}

In order to show the efficacy of our design, we run our system with LH+MH+SH module combination in an eight hours long wall time (six hours long simulation time) simulation \emph{without any collisions and deadlocks}.
The environment is forest-like and contains $300$ randomly generated dynamic obstacles in which eight quadrotors navigate.
The recording be found in the long term execution supplemental video.\looseness=-1

\section{Conclusion}

In this paper, we propose MRNAV, a multi-robot aware planning and control stack design for collision and deadlock-free navigation in cluttered environments with static and dynamic obstacles.
Our design utilizes three qualitatively different decision making modules, that are often seen as alternatives for collision avoidance in conventional designs.
We assign responsibilities to the decision making modules, arguing that they complement each other.
MRNAV is the first hierarchical approach combining these three qualilatively different decision making modules. 
We define the abstractions we require for the interfaces to the perception, localization, mapping, and prediction subsystems.
We implement our design for simulated multi-quadrotor flight, and discuss our algorithm choices and integration of the modules to the simulation environment.
In our evaluations, we show that all three decision making modules are required for effective navigation in diverse types of environments.
We show the long-term executability of our approach in an eight hour uninterrupted run without collisions or deadlocks.
Future work includes evaluating the navigation stack in real-world experiments.\looseness=-1


\bibliographystyle{ieeetr}
\bibliography{bibliography.bib}

\end{document}